\documentclass[11pt,a4paper]{article}
\usepackage[hyperref]{acl2021}
\usepackage{times}
\usepackage{latexsym}

\usepackage{microtype}

\aclfinalcopy 

\RequirePackage{amsmath}
\usepackage{multirow}
\usepackage{comment}
\usepackage{graphicx}
\usepackage{subcaption}

\usepackage{fancyhdr}
\title{Regressing Location on Text for Probabilistic Geocoding}

\author{Benjamin J. Radford \\
  University of North Carolina at Charlotte\\
  \texttt{benjamin.radford@uncc.edu} \\}

\date{}

\begin{document}
\maketitle
\thispagestyle{fancy}
\begin{abstract}
Text data are an important source of detailed information about social and political events. Automated systems parse large volumes of text data to infer or extract structured information that describes actors, actions, dates, times, and locations. One of these sub-tasks is geocoding: predicting the geographic coordinates associated with events or locations described by a given text. We present an end-to-end probabilistic model for geocoding text data. Additionally, we collect a novel data set for evaluating the performance of geocoding systems. We compare the model-based solution, called ELECTRo-map, to the current state-of-the-art open source system for geocoding texts for event data. Finally, we discuss the benefits of end-to-end model-based geocoding, including principled uncertainty estimation and the ability of these models to leverage contextual information. 
\end{abstract}

\section{Introduction}
\label{sec:overview}

Text data are an important source of information about social and political events. We introduce a novel method for predicting the latitude and longitude of locations mentioned or described in natural language texts (``geocoding''). This neural network-based method offers several advantages over existing rule-based techniques for geocoding: (1) it produces a probability distribution over predicted latitudes and longitudes thereby allowing users to report the certainty of their estimates; (2) it does not require the identification of place names in the text prior to geocoding; (3) it naturally leverages contextual clues to improve predictions and disambiguate location names.

This paper proceeds by first providing a brief overview of related work in geocoding and language modeling. We then introduce a probabilistic model for geocoding texts and identify a dataset with which to train and evaluate the model. We compare our results to existing methods and conclude with suggestions for future research.


\subsection{Geocoding Text}\label{sec:geocoding}

\citet{lee:liu:ward:2019} describe a geolocation pipeline for producing political event data that includes three steps: \newcounter{foo} \refstepcounter{foo}(\thefoo)\label{step:ner} named entity recognition (NER) identifies character strings of named places; \refstepcounter{foo}(\thefoo)\label{step:geoparsing} ``geoparsing'' software matches named locations to geographical locations; \refstepcounter{foo}(\thefoo)\label{step:linking} events from the source text are linked to their respective locations.

Mordecai is an open source tool for Steps~\ref{step:ner} and \ref{step:geoparsing} \citep{halterman:2017}. Mordecai uses a pretrained named entity recognition model and word2vec \citep{mikolov:etal:2013} to match location names identified within an unstructured text document to known locations within the GeoNames Gazetteer \citep{geonames}. 

\citet{kulkarni:etal:2020} present a model-based geocoding solution. Their convolutional neural network model predicts geographic grid cell membership for each input text; it does not predict latitude and longitude values directly. This complicates comparison with the model presented here which directly regresses latitude and longitude on text. For example, the evaluation metrics the authors chose for their model are largely based on classification accuracy rather than continuous measures of nearness, as would be the case in a regression setting.\footnote{Specifically, the authors report the area under the receiver operating characteristic curve (AUC) and classification accuracy. This classification framing contrasts with the model presented here which directly predicts latitude and longitude values and therefore is evaluated via mean absolute error in kilometers.} 

\subsection{Transformer Language Models}

The foundation of the model described in this paper is a very large neural network language model called a transformer network, a ``transformer.'' Typically, a transformer is trained on a large corpus with a self-supervised objective: either next sentence prediction and/or masked language prediction. This initial training is called ``pretraining.''  However, these models have been shown to generalize very well to tasks for which they were not explicitly pretrained. With subsequent ``fine-tuning,'' transformers can acquire the ability to accomplish new tasks with substantially fewer training examples than those with which they were pretrained. \citet{vaswani:etal:2017} introduced the first transformer language model; the particular model used here is called DistilRoBERTa \citep{sanh:etal:2019, liu:etal:2019}.

\section{Model}

We introduce a model that is capable of performing Steps~\ref{step:ner} through \ref{step:linking} (\S~\ref{sec:geocoding}) end-to-end. That is, given training data exemplary of the desired mapping from text inputs, $\textbf{X}$, to geographic coordinates, $\textbf{Y}$, this model is fine-tuned such that it learns a function $f(\textbf{x}_i; \textbf{W}) \rightarrow \hat{\textbf{y}}_i$, where $\textbf{W}$ is the set of model parameters. This is a non-linear multivariate regression of latitude and longitude on text. We modify a pretrained DistilRoBERTa model by adding three fully-connected dense layers with sigmoid activation, an output (``head'') layer, and a custom loss function. We use this model to minimize the negative log likelihood of a five component mixture of von Mises-Fisher (vMF) distributions conditional on the input text.

\begin{table*}
    \centering
    \begin{tabular}{lrrrrrrr}
        \hline
        & & \multicolumn{2}{c}{\textbf{\emph{highProb}}} & \multicolumn{2}{c}{\textbf{\emph{best}}} & \multicolumn{2}{c}{\textbf{\emph{random}}} \\ \cline{3-4} \cline{5-6} \cline{7-8}
        \textbf{Model} & \multicolumn{1}{c}{$\textbf{n}$} & \multicolumn{1}{c}{Mean} & \multicolumn{1}{c}{Median} & \multicolumn{1}{c}{Mean} & \multicolumn{1}{c}{Median} & \multicolumn{1}{c}{Mean} & \multicolumn{1}{c}{Median} \\ \hline
        Mordecai & 12,864 & 1101.3 & 161.9 & 348.8 & 14.3 & 1213.7 & 140.6  \\
        & & (24.1) & (2.9) & (13.5) & (0.4) & (24.0) & (3.5) \\
        Mordecai Complete Cases & 12,585 & 946.0 & 154.5 & 177.0 & \textbf{13.4} & \textbf{1076.7} & \textbf{134.0}   \\
        & & (22.0) & (3.1) & (8.3) & (0.3) & (25.0) & (4.2) \\
        ELECTRo-map & 12,864 & \textbf{108.1} & \textbf{44.1} & \textbf{96.8} & 44.0 & 6380.8 & 4814.5  \\
        & & (4.1) & (0.4) & (2.5) & (0.4) & (54.1) & (96.1) \\
        \hline
    \end{tabular}\caption{Test set (out-of-sample) geocoding performance. Reported values are measured in kilometers. Bootstrap estimated standard errors in parentheses.}\label{tab:evaluation}
\end{table*}

\begin{figure*}
        \centering
        \begin{subfigure}[b]{0.475\textwidth}
            \centering
            \includegraphics[width=\textwidth]{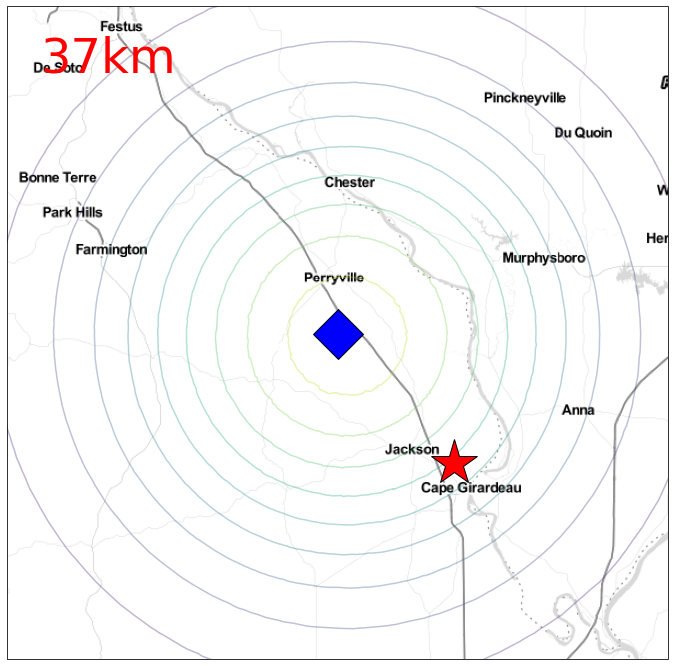}
            \caption[]%
            {{\small ``Hanover Lutheran Church is a Lutheran congregation in Cape Girardeau, Missouri, that is a member of the Lutheran Church–Missouri Synod...''}}    
            \label{fig:a}
        \end{subfigure}
        \hfill
        \begin{subfigure}[b]{0.475\textwidth}  
            \centering 
            \includegraphics[width=\textwidth]{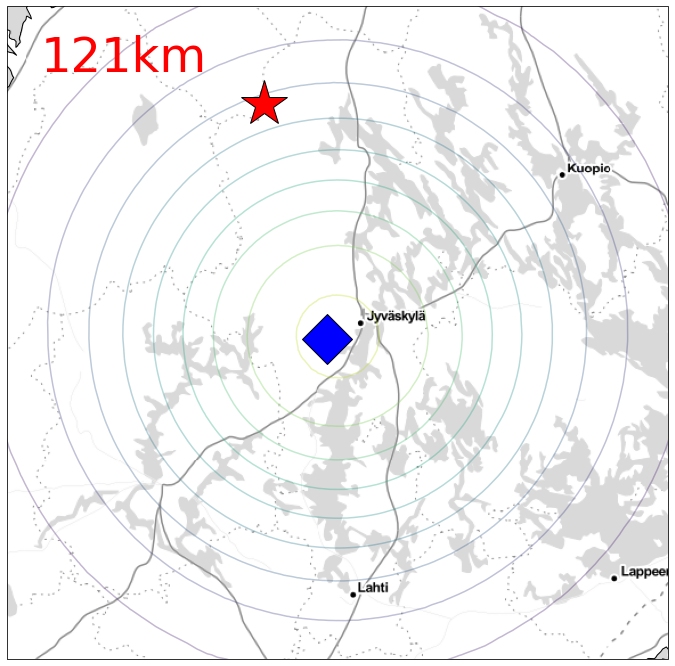}
            \caption[]%
            {{\small ``Salamanperä Strict Nature Reserve (Salamanperän luonnonpuisto) is home to Wolverine and Finnish Forest Reindeer (R. tarandus fennicus), and it is said...''}}    
            \label{fig:b}
        \end{subfigure}
        \vskip\baselineskip
        \begin{subfigure}[b]{0.475\textwidth}   
            \centering 
            \includegraphics[width=\textwidth]{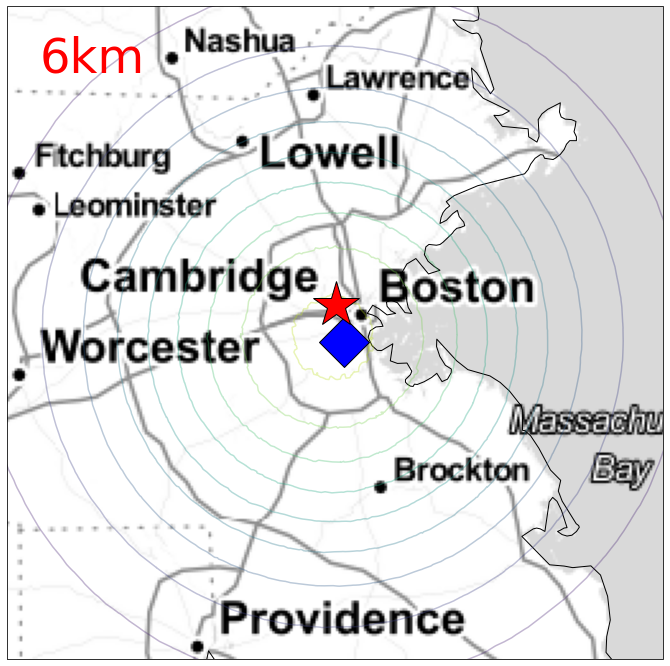}
            \caption[]%
            {{\small ``Houghton Library, on the south side of Harvard Yard adjacent to Widener Library, is Harvard University's primary repository for rare books and manuscripts...''}}    
            \label{fig:c}
        \end{subfigure}
        \hfill
        \begin{subfigure}[b]{0.475\textwidth}   
            \centering 
            \includegraphics[width=\textwidth]{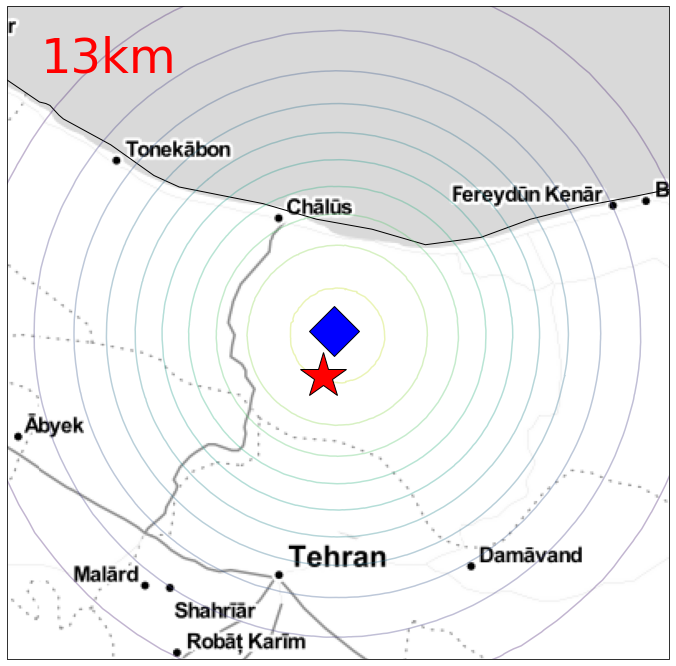}
            \caption[]%
            {{\small ``Pil (Persian --; also Romanized as Pīl; also known as Pel) is a village in Owzrud Rural District, Baladeh District, Nur County, Mazandaran Province, Iran...''}}    
            \label{fig:d}
        \end{subfigure}
        \caption[ The average and standard deviation of critical parameters ]
        {Predicted (diamond) and actual (star) coordinates. Contours represents 10\% of the vMF mixture probability density. Approximately 95\% of the probability density for each vMF mixture is shown in Figures~\ref{fig:a}--\ref{fig:d}.} 
        \label{fig:maps}
    \end{figure*}

The von Mises distribution is an approximation of a univariate Gaussian distribution on the circumference of a circle. The vMF distribution generalizes the von Mises distribution beyond two dimensions to the surfaces of spheres and hyperspheres; when $p=2$, the vMF distribution is equivalent to the the von Mises distribution.

Because the vMF distribution has support over the surface of the unit $p-1$ sphere in $p$ Euclidean space, we must transform our geodetic coordinates (latitude and longitude) to Cartesian coordinates on this sphere. The formulae to do so, assuming a spherical Earth, are given by Equations~\ref{eq:xi}--\ref{eq:zi}.
\begin{align}
    x_i &= \text{cos} ( \text{rad}^{lat}_i ) \times \text{cos} ( \text{rad}^{lon}_i ) \label{eq:xi} \\ 
    y_i &= \text{cos} ( \text{rad}^{lat}_i ) \times \text{sin} ( \text{rad}^{lon}_i ) \label{eq:yi}\\
    z_i &= \text{sin} ( \text{rad}^{lat}_i ) \label{eq:zi}
\end{align}

The vMF probability density function is given by Equation~\ref{eq:vmf}. $\mu$, the mean direction, is a point in $p$ space that falls on the unit $p-1$ sphere. A point $x$ in $p$ space can be projected onto this sphere by $L2$ normalization: $x/||x||$. The concentration parameter, $\kappa$, controls the dispersion of the distribution across the surface of the sphere. $\kappa=0$ corresponds to a uniform distribution over the entire sphere while $\kappa=\infty$ corresponds to a point mass at $\mu$. $I_{p/2-1}$ is the modified Bessel function of the first kind at order $p/2-1$.
\begin{align}\label{eq:vmf}
    f_{vmf}(x;\mu,\kappa) &=  \frac{\kappa^{p/2-1}}{(2\pi)^{p/2}I_{p/2-1}(\kappa)}e^{(\kappa \mu^T x)}
\end{align}

A probabilistic neural network model with a single vMF component is optimized by minimizing the negative log likelihood given in Equation~\ref{eq:nll}.
\begin{align}\label{eq:nll}
    -L(\textbf{W}) &= -\sum_{i} \text{ln}(f_{vmf}(\textbf{y}_i;\hat{\mu}_i,\hat{\kappa}_i)) \\ \notag
    \hat{\mu}_i &= f_{\mu}(\textbf{x}_i; \textbf{W}) \\ \notag
    \hat{\kappa}_i &= f_{\kappa}(\textbf{x}_i; \textbf{W})
\end{align}

The outputs of the neural network, given an input text $\textbf{x}_i$, are the parameters of a vMF distribution. Therefore, the model estimates a distribution over possible coordinates for a given input text. While the parameters of the neural network itself ($\textbf{W}$) are deterministic, predicting a probability distribution for each input text allows us to capture aleatoric uncertainty. Aleatoric uncertainty is the uncertainty inherent in the data themselves. In the case of geocoding text, this uncertainty may result from texts that do not distinguish between Springfield, IL and Springfield, GA, or from texts that refer to multiple locations (assuming that the model in question is unable to represent a multimodal distribution). 

This uncertainty is unlikely to be homoskedastic; some texts will more precisely specify relevant locations than others. We allow for heteroskedastic uncertainty by estimating both the central tendency ($\hat{\mu}_i$) and the dispersion ($\hat{\kappa}_i$) of a target distribution.

Building on the negative vMF log likelihood loss described above, we optimize a neural network model to predict a mixture of vMF distributions.\footnote{We use the Adam optimizer with a learning rate of $5\times10^{-5}$ and train for five epochs \citep{DBLP:journals/corr/KingmaB14}. The network is difficult to train and a single-component vMF model failed to converge.} For every input text, the model predicts parameters for five vMF distributions in addition to a set of mixing probabilities. The mixing probabilities describe the weights associated with each of the five vMF components. In this way, the model can fit a more flexible distributional shape than it would otherwise be able to with a single vMF component. Indexing the mixing components, $\rho$, by $k$, the revised loss function is given in Equation~\ref{eq:mixloss}. We refer to this model as ELECTRo-map: End-to-end Location Estimation with Confidence via Transformer Regression.\footnote{\url{https://tfwiki.net/wiki/Electro_map}}
\begin{align}\label{eq:mixloss}
    -L(\textbf{W}) &= - \sum_{k=1}^{5} \rho_k \sum_{i} \text{ln}(f_{vmf}(\textbf{y}_i;\hat{\mu}_{ik},\hat{\kappa}_{ik}))
\end{align}

\section{Data}

To evaluate ELECTRo-map, we collect data from all Wikipedia articles with coordinates linked to Wikidata.org.\footnote{Found at \url{https://www.wikidata.org/wiki/Q15181105}} These data include the primary latitude and longitude associated with an article, globe, title, language, and extract attributes. The data were collected via the official Wikipedia API by iterating over the set of Wikipedia pages linked to Wikidata geographic entries.\footnote{\url{https://en.wikipedia.org/w/api.php}}  Together, the data comprise the introductory sections of 1,286,475 English language articles. Most of the excerpts are between one sentence and a couple paragraphs in length. Many of these texts contain references to multiple geographic locations, but each one only has one ``correct'' latitude and longitude pair that describes the precise location of the article's referent. These are partitioned into a training set (1,260,746 articles), a validation set (12,864 articles) and a test set (12,865 articles).\footnote{Test set size is kept small due to hardware limitations and the speed of Mordecai.}

\section{Evaluation}

We compare the performance of ELECTRo-map against Mordecai. Because Mordecai and ELECTRo-map can both return multiple results per text, we offer three solutions for aggregating results to a single latitude and longitude prediction per observation. The first is to take the single highest probability prediction (\textit{highProb}).\footnote{While ELECTRo-map produces proper probabilities for each component, Mordecai only produces a country-level confidence score.} The second is to take the best prediction from the mixture (\textit{best}).\footnote{Note that this rule requires knowledge of the target latitude and longitude. It therefore represents an unrealistic ideal scenario.} The third is to take a random prediction from the mixture (\textit{random}). Mordecai occasionally returns null results. In these cases, we impute a latitude and longitude pair of (0.0, 0.0). We also provide results for a complete cases analysis of Mordecai, omitting all 279 observations for which Mordecai failed to produce a geolocation.

Results are shown in Table~\ref{tab:evaluation}. In the best case scenario, that in which the location of interest is known a priori, Mordecai clearly outperforms ELECTRo-map. Mordecai's median error is only 13.4km. However, in the more likely scenario that a single geolocation is desired for a text and no a priori knowledge of the preferred prediction is available, ELECTRo-map outperforms Mordecai. Mean and median errors for ELECTRo-map are 108.1km and 44.1km, respectively, compared to 946km and 154.5km for Mordecai. These numbers also compare favorably to the \citet{kulkarni:etal:2020} model; in addition to classification-based metrics, the authors report the mean distance between predicted grid cell centroids and target locations. They report mean errors of between 174km and 180km.\footnote{Note that the data sets in these two papers, while both based on Wikipedia, are distinct.}

Four examples drawn from the test set are depicted in Figure~\ref{fig:maps}. Predicted and actual locations are given as well as contours denoting the probability density associated with the predicted distribution. Each contour represents one decile. Each subfigure represents roughly 95\% of the probability density. Captions give abridged excerpts of the associated input texts.

\section{Conclusion}

When humans perform geocoding manually, they often rely on contextual clues for assistance. Those clues may or may not come from the text itself. For instance, the presence of other named entities, like sports teams, may help human coders to distinguish between Washington state and Washington D.C. Automated processes for geocoding should also make use of contextual clues.

Model-based geocoding offers a natural method for both incorporating contextual clues and for dealing with the uncertainties that arise while geocoding. ELECTRo-map, for instance, quantifies uncertainty by estimating a mixture of probability distributions over likely geographic coordinates. Furthermore, model-based geocoding offers the ability to fine-tune for specific tasks: researchers may be interested in geocoding certain parts of texts and not others (e.g. birth and death places). To the extent that the model is unable to distinguish between multiple location types in the source text, this ambiguity should be reflected in the model's reported uncertainty. Model-based and gazetteer-based methods (like Mordecai) are not exclusive, though. It may be possible to derive better results by, for example, first identifying a distribution over likely locations via a statistical model and then ``snapping to'' a most likely location within that distribution using a gazetteer.

Finally, the success of multilingual transformers suggests that ELECTRo-map or related techniques may generalize across languages \citep{K2020Cross-Lingual}. Future efforts on model-based geocoding should seek to evaluate cross-lingual performance and measure the importance of context on location disambiguation.

\section*{Acknowledgments}

We thank Dr. Yaoyao Dai of UNC Charlotte and three anonymous reviewers for their helpful feedback and suggestions.

\bibliographystyle{acl_natbib}
\bibliography{refs,anthology,acl2021}


\end{document}